\pdfoutput=1

\documentclass[11pt]{article}

\usepackage{ACL2023}

\usepackage{times}
\usepackage{latexsym}

\usepackage[T1]{fontenc}

\usepackage[utf8]{inputenc}

\usepackage{microtype}

\usepackage{inconsolata}
\usepackage{graphicx}
\usepackage{amsmath}
\usepackage{adjustbox}
\usepackage{multirow}
\usepackage{svg}
\usepackage[normalem]{ulem}
\useunder{\uline}{\ul}{}
\usepackage{makecell}

\usepackage{xcolor}

\definecolor{colorhigh}{HTML}{9900FF}
\definecolor{colorlow}{HTML}{660000}

\title{BIOptimus: Pre-training an Optimal Biomedical Language Model with Curriculum Learning for Named Entity Recognition}

\author{Vera Pavlova \\
  rttl.ai \\ Dubai, UAE \\
  {v@rttl.ai} \\\And
  Mohammed Makhlouf \\
  rttl.ai \\ Dubai, UAE \\
  {mm@rttl.ai} \\}

\begin{document}
\maketitle
\begin{abstract}
Using language models (LMs) pre-trained in a self-supervised setting on large corpora and then fine-tuning for a downstream task has helped to deal with the problem of limited label data for supervised learning tasks such as Named Entity Recognition (NER). Recent research in biomedical language processing has offered a number of biomedical LMs pre-trained using different methods and techniques that advance results on many BioNLP tasks, including NER. However, there is still a lack of a comprehensive comparison of pre-training approaches that would work more optimally in the biomedical domain. This paper aims to investigate different pre-training methods, such as pre-training the biomedical LM from scratch and pre-training it in a continued fashion. We compare existing methods with our proposed pre-training method of initializing weights for new tokens by distilling existing weights from the BERT model inside the context where the tokens were found. The method helps to speed up the pre-training stage and improve performance on NER. In addition, we compare how masking rate, corruption strategy, and masking strategies impact the performance of the biomedical LM. Finally, using the insights from our experiments, we introduce a new biomedical LM (BIOptimus), which is pre-trained using Curriculum Learning (CL) and contextualized weight distillation method. Our model sets new states of the art on several biomedical Named Entity Recognition (NER) tasks. We release our code and all pre-trained models.\footnote{The source code and the models are released on our Github \href{https://github.com/rttl-ai/BIOptimus}{https://github.com/rttl-ai/BIOptimus}, including the Rust code for qik-find.}
\end{abstract}

\begin{figure}[h]
\centering
\includegraphics[width=0.9\columnwidth]{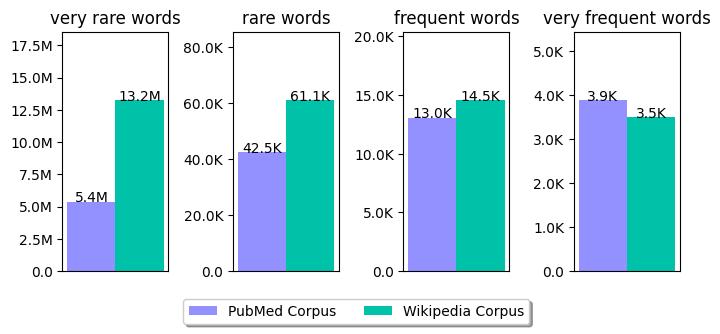}
\caption{Comparison of word frequency (raw count) for Wikipedia Corpus and PubMed corpus. Frequencies are divided into four categories. Each bar chart represents how many words of that frequency category are found in both corpora.}
\end{figure}

\section{Introduction}

Since the introduction of transformer architecture \cite{10_vaswani_2017}, transfer learning has gained immense popularity in Natural Language Processing (NLP). Pre-training LMs on a large corpus (\citealp{11_devlin_2019}; \citealp{12_liu_2019}) allows a large-scale knowledge extraction. In the current pipeline of knowledge transfer in NLP, the pre-training stage can be viewed as preparing the model with broad general knowledge \cite{2_raffel_2019} that can later be fine-tuned to new tasks or adapted to different domains. However, applying these models in a new domain directly without additional pre-training steps cannot achieve high performance because of a word distribution shift from a general domain vocabulary to a new domain vocabulary. The importance of a pre-trained model for the biomedical domain is even more significant due to the difficulties of constructing supervised training data because it requires expert knowledge for quality annotations.

The biomedical domain has greatly benefitted from domain-specified LMs. To date, various methods have been developed to pre-train biomedical language models: pre-training from scratch (\citealp{32_gu_2021}; \citealp{5_beltagy_2019}), continual pre-training (\citealp{6_lee_2019}; \citealp{53_huang_2019}), and hybrid approaches (\citealp{tai_2020}; \citealp{35_poerner_2020}; \citealp{34_sachidananda_2021}). Nonetheless, more research is needed to compare the optimal ways to pre-train the biomedical LMs. Comparison of different pre-training approaches and techniques primarily targets general domain LMs (\citealp{wettig2023mask}; \citealp{dai-etal-2022-whole}; \citealp{yamaguchi-etal-2021-frustratingly}; \citealp{gu-etal-2020-train}). What still needs to be clarified is the effects of these methods on the pre-training biomedical LM.

Typically, general domain LMs are pre-trained on Wikipedia, Book Corpora, and Common-Crawl (\citealp{11_devlin_2019}; \citealp{lan2020albert}; \citealp{25_radford_2019}; \citealp{2_raffel_2019}; \citealp{16_clark_2020}). These corpora are considerably different from the biomedical corpora in terms of domain specifics (use of biomedical terminology) and the style of scientific text (most of the models are trained on PubMed). Thus, the methods and techniques that work well for the general domain are not guaranteed to work the same for the biomedical domain.
Figure 1 illustrates the difference in the raw counts of word frequencies between Wikipedia and PubMed, which contains only abstracts (PubMed Corpus). The corpora are of similar sizes (3.1 and 3.6 billion tokens, correspondingly). We can observe that, for example, the count of rare words noticeably differs between two corpora and twice less for the PubMed Corpus. \citet{yu-etal-2022-rare} and \citet{46_ethayarajh_2019} showed how the presence of rare words in the pre-training corpus might greatly impact the performance of the LM. 
This work aims to compare different techniques and methods for creating biomedical LMs. We compare such approaches as pre-training from scratch, continued pre-training with no specialized vocabulary, and continued pre-training with biomedical vocabulary. Our evaluations show that a biomedical LM pre-trained in a continued setting speeds up the pre-training. To account for the lack of specialized vocabulary that inevitably accompanies models pre-trained following a continued approach, we propose a new method to initialize weights for new tokens by distilling existing weights from the BERT model inside the context where they were found. Furthermore, we compare how masking rate, corruption strategy, and masking strategies impact the performance of the biomedical LM. We observe that the corruption strategy of "80-10-10" \cite{11_devlin_2019} may play a role in the representation degeneration problem of LM by increasing  the degree of anisotropy in the contextualized word representations. With relation to the word representation, anisotropy characterizes an embedding space that is restricted into a narrow cone-like shape \cite{46_ethayarajh_2019}. This phenomenon leads to a decrease in word embeddings' expressiveness and is referred to as the representation degeneration problem \cite{DBLP:journals/corr/abs-1907-12009}.

Finally, based on the results of our experiments, we introduce a model which pre-trained with a CL schedule. CL is an easy-to-hard training strategy \cite{soviany2022curriculum}. Though very successful and widely adopted in NLP, it is still under-explored in pre-training LMs. To the best of our knowledge, CL was not used before to pre-train the biomedical LM. We propose the CL method based on the task complexity of predicting masked labels. Our model (BIOptimus), pre-trained with the contextualized weight distillation and following our CL method, sets new states of the art on several biomedical Named Entity Recognition (NER) tasks.
In summary, our contributions are:

\begin{itemize}
\item Firstly, we propose a new approach to initialize weights to pre-train a biomedical LM that leverages the efficiency of pre-training in a continued fashion with specialized biomedical vocabulary that improves the model's performance on NER tasks.

\item Secondly, we comprehensively compare our approach with other pre-training approaches. In addition, we experiment with masking ratio, corruption strategy, and masking strategies, testing what works more optimally to prepare a language model for the biomedical domain.

\item Following experimental insights, we propose a biomedical model (BIOptimus) pre-trained with the CL method and contextualized weight distillation.
\end{itemize}

\section{Prior Literature}
Recent studies indicate that adapting existing LMs pre-trained on general corpora for a new domain or pre-training a new LM from scratch for a specific domain gives a substantial gain in terms of performance for a downstream task. The authors of BioBERT \cite{6_lee_2019} performed domain adaptation by continuing pre-training BERT \cite{11_devlin_2019} on a large corpus of biomedical data. The model showed notable achievement on downstream tasks, outperforming BERT and the state-of-the-art models. \citet{53_huang_2019} further pre-train BioBERT using clinical corpora to address clinical text's challenges and released ClinicalBERT. \citet{7_gururangan_2020} showed the benefits of continued domain pre-training followed by task-adaptive pre-training (TAPT) of RoBERTa on four domains, including biomedical.
One of the downsides of continued pre-training is the lack of domain-specific vocabulary. The authors of the SciBERT model \cite{5_beltagy_2019} constructed a new Scivocab using the SentencePiece library, introducing 58\% of new tokens, which shows how much scientific vocabulary differs from a general vocabulary of BERT-base. The pre-training corpus comprises 82\% of the biomedical domain; the rest is the computer science domain. Training from scratch solely on the biomedical domain showed further gain and was presented by the authors of PubMedBert \cite{32_gu_2021}. They construct a new domain-specific vocabulary, proving its effectiveness when fine-tuning for downstream tasks. Another approach to pre-train the biomedical LM from scratch using citation links is the LinkBERT model, which achieved new state-of-the-art on several BioNLP tasks \cite{yasunaga-etal-2022-linkbert}. 
To mitigate the high costs of pre-training from scratch, \citet{tai_2020} introduced extension vocabulary that is not found in the vocabulary of BERT. The weights of the embedding layer of the extension vocabulary are randomly initialized, then further pre-trained on the biomedical corpus. To avoid randomly initializing weights for new tokens and speed up pre-training, \citet{35_poerner_2020} used word2vec. Word2vec vectors are trained on the biomedical domain and then aligned with wordpiece vectors from BERT. \citet{34_sachidananda_2021} experimented with subword-based initialization using the mean of RoBERTa fixed subword embeddings.

\textbf{Curriculum Learning}
Curriculum learning (\citealp{ELMAN199371}; \citealp{bengio_2009}) has been successfully explored in NLP to solve different tasks. It was widely adopted in machine translation (\citealp{Guo_Tan_Xu_Qin_Chen_Liu_2020}; \citealp{liu2020a}; \citealp{wang-etal-2020-learning-multi}, \citealp{Zhan_Liu_Wong_Chao_2021}). Some works have been done in the area of answer generation \cite{liu_2018}, relation extraction \cite{huang-du-2019-self}, reading comprehension \cite{tay2019simple}, and NER \cite{jafarpour-etal-2021-active}. \citet{xu-etal-2020-curriculum} used CL during the fine-tuning stage of BERT. \citet{nagatsuka-etal-2021-pre} applied CL to pre-train RoBERTa by gradually increasing the block size of the text. \citet{lee-etal-2022-efficient-pre} proposed a concept-based curriculum masking.

\section{Method}

In continued biomedical domain adaptation, pre-training using the checkpoint of the model pre-trained on the general domain can significantly speed up training. Moreover, the benefits of continued domain adaptation are highly aligned with the present trends of NLP communities \cite{1_bommasan_2021} of reducing the environmental impact and making knowledge readily available in limited resources settings. One of the problems of continued pre-training of LMs for domain adaptation is, loosely speaking, the absence of domain-specific vocabulary \cite{33_guo_2022}. The issue of out-of-vocabulary (OOV) words is more elegantly solved with subword tokenization such as Byte-Pair Encoding (BPE) \cite{sennrich-etal-2016-neural}, WordPiece \cite{37_song_2021}, and SentencePiece \cite{kudo-richardson-2018-sentencepiece}. Nevertheless, the presence of a whole word or more meaningful subwords in domain-specific vocabulary greatly enhances performance on downstream tasks (\citealp{5_beltagy_2019}; \citealp{32_gu_2021}; \citealp{33_guo_2022}). There are ways to address this issue by incorporating domain-specific tokens and assigning new token weights without initializing them from scratch (\citealp{34_sachidananda_2021}; \citealp{35_poerner_2020}). Nevertheless, none of the current approaches leverage the advantage of contextualized embeddings of transformer models. To boost knowledge transfer, we introduce the new vocabulary by distilling existing weights from the BERT model inside the context of the domain where they were found.
Our approach uses WordPiece Tokenization \cite{37_song_2021} to tokenize PubMed Corpus \footnote{\url{https://pubmed.ncbi.nlm.nih.gov}. Abstracts published before Jan 2023.} (BioMedTokenizer) with a vocabulary size of 30522.  We contextualize tokens in the following way (see Figure 2): 

\begin{itemize}
    \item First, we perform tokenization with BioMedTokenizer.  For example, suppose BioMedTokenizer has a token "bronchoconstriction" absent from the original bert-base-uncased vocabulary. In that case, the token will be broken into six pieces: ['bro','\#\#nch','\#\#oco','\#\#nst','\#\#ric', '\#\#tion'] (see Figure 2). We use the mean operation of constituting tokens to compute a single representation for a new token (distilled representations):
    \begin{equation}
    \begin{split}
        t_{distilled} = f(t_1,...,t_k) \\ f \in \{mean\}\\
    \end{split}
    \end{equation}
    The variable \emph{k} represents the number of the subtokens that make up the token of interest.
    We average token weights of only the last layer of the BERT model as it is more expressive of context and domain information \cite{29_peters_2018}. 

    \item  In the next step, we compute aggregated representation by sampling sentences that contain tokens of interest from the same PubMed Corpus and compute aggregated weight across several sentences (contextualized representations):
    \begin{equation}
    \begin{split}
        t_{context} = g(t_{distilled},...,t_m) \\ g \in \{mean\}\\
    \end{split}
    \end{equation}
     The size of sentences sampled per token \emph{m} may vary. We sample uniformly at random, setting the upper bound equal to 20 and the lower bound to 1. In case the token is not found in the corpus (such tokens represent less than 1/10 of the whole vocabulary), we assign distilled representation to this token. We use mean operation, which was found to be the most efficient operation for aggregating across several contexts \cite{bommasani_2020}. 
\end{itemize}

The averaging of the weights of subtokens constitutes more meaning when placed in context. Moreover, contextualization of tokens' embeddings leverages the ability of the BERT model to create domain-specific word embeddings that align with the corpus where they were found.  \cite{13_aharoni_goldberg_2020}. We call this approach \textbf{contextualized weight distillation}.
\begin{figure*}[h]
\centering
\includegraphics[width=\textwidth]{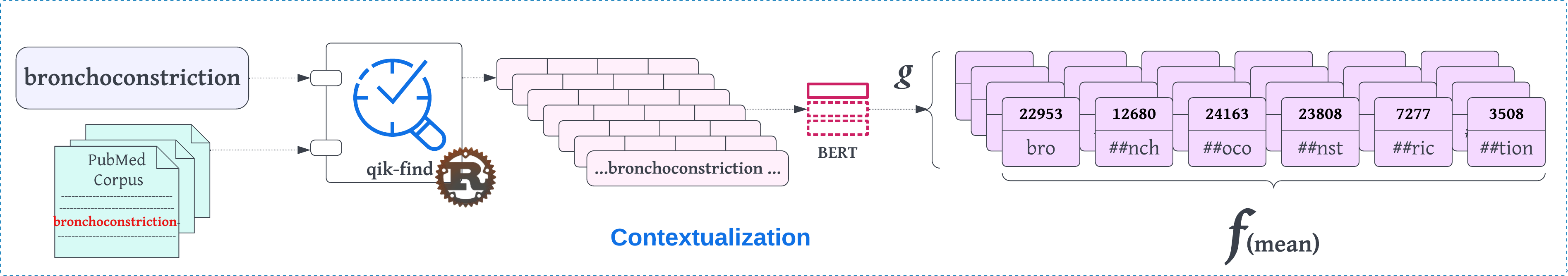}
\caption{Contextualization of tokens' embeddings with our resource-efficient and performant qik-find tool written in the Rust programming language. Qik-find is a purpose-built tool to find tokens of interest and extract corresponding sentences from a large corpus exploiting the native capability of the Rust programming language for efficient multiprocessing.}
\end{figure*}

\section{Models}

To perform a more rigorous comparison and find an optimal approach to the pre-train biomedical LM, we pre-train four models (see Table 1):
\begin{itemize}
\item The first model is pre-trained from scratch, further referred to \textbf{as the biomedical model from scratch (BM from scratch)}. We train WordPiece tokenizer on PubMed corpus and construct new biomedical vocabulary. The model is trained on the same corpus with all the weights initialized randomly.
\item The second model is pre-trained from the BERT-base model checkpoint\footnote{\url{https://huggingface.co/bert-base-uncased}}.  We call it \textbf{the biomedical model continued (BM continued)}.
\item The third model is pre-trained in a hybrid approach called \textbf{the biomedical model averaged token weights (BM averaged)}. We use BioMedTokenizer to construct the biomedical vocabulary and find an intersection between BERT-base vocabulary and biomedical vocabulary. The weights of the tokens, which are common for both vocabularies, are copied from BERT-base vocabulary to new model weights by directly mapping weights to the corresponding tokens or subtokens. The weights of new domain-specific tokens absent in the BERT-base model are synthesized by averaging the corresponding subtokens representations and assigning this averaged representation to the token of interest in the embedding matrix of the new model. This method corresponds to the step "distilled representations" described in Section 3.
\item The fourth model is pre-trained following the proposed method in Section 3. We refer to this approach \textbf{biomedical model contextualized weights (BM contextualized)} or \textbf{BIOptimus 0.1}. We assign tokens' weights that are common for BERT-base vocabulary and newly constructed biomedical vocabulary in the same manner as described above in the \textbf{biomedical model averaged token weights}. Contextualization of tokens is performed only for the tokens that are not found in the BERT-base vocabulary.
\end{itemize}

\begin{table}[!htbp]

\begin{adjustbox}{width=\columnwidth,center}
\begin{tabular}{c|c|c}
\hline
\multicolumn{1}{l|}{}                            & \textbf{\thead{Pre-training\\method}} & \textbf{\thead{Weight\\initialization}}                                    \\ \hline
\textbf{\thead{BM\\from scratch}}           & from scratch                 & randomly                                                          \\ \hline
\textbf{\thead{BM\\continued}}              & continued                    & from the existing checkpoint                                      \\ \hline
\textbf{\thead{BM\\averaged}} & continued                    & \begin{tabular}[c]{@{}c@{}}from the existing checkpoint \\+ averaging  subtokens\end{tabular}               \\ \hline
\textbf{\thead{BM\\contextualized}}         & continued                    & \begin{tabular}[c]{@{}c@{}}from the existing checkpoint \\+ contextualized weight distillation\end{tabular} \\ \hline

\end{tabular}
\end{adjustbox}
\caption{Four pre-trained models with different pre-training approaches and initialization methods.}
\label{tab:my-table}
\end{table}

\section{Data}
The primary data used for pre-training is the same PubMed Corpus used for training BioMedTokenizer. To speed up experiments, we choose a random subset of this corpus of 1.8 billion words for pre-training models, representing approximately half of the data generally used for pre-training models for biomedical domain adaptation. 
To test our models on the downstream task, we use NER tasks. NER is one of the most fundamental biomedical tasks; it is an essential first step in processing literature for biomedical text mining. Moreover, the NER is an excellent test for evaluating a domain-specific model's success to recognize different types of biomedical terminology. The recent initiative of Microsoft Research \cite{32_gu_2021} unified and released a new BLURB benchmark (Biomedical Language Understanding Reasoning Benchmark). The Benchmark includes five NER datasets: BC5-chem \cite{li_2016}, BC5-disease \cite{li_2016}, NCBI-disease \cite{dogan_2014}, BC2GM \cite{smith_2008}, JNLPBA \cite{kim_2004}. Label distribution \cite{crichton_2017} is presented in Appendix A.

\section{Experimental Set-up}

We pre-train all models with the same hyperparameters (details can be found in Appendix A). We further fine-tune each model for five NER datasets and report results in Table 2. Fine-tuning details can be found in Appendix A. To compare the effectiveness of pre-training approaches more closely, we show the effects of each pre-training epoch on the performance of our models on three datasets: JNLPBA, BC5-chem, and NCBI-disease (Figure 3). We use the same evaluation metric and fine-tuning approach for the NER task described in PubMedBERT \cite{32_gu_2021}.

Moreover, we experiment with other pre-training techniques. The above-described models are pre-trained by masking random tokens. \citet{32_gu_2021}, and \citet{Cui_2021} showed that pre-training with whole-word masking (WWM) can be more beneficial. \citet{wettig2023mask} experimented with different percentages of masking rate and found out that for the BERT-base, 0.2 percent may be more optimal to prompt the model to learn better. Thus, we pre-train two more models using our method but with different techniques, one model using WWM and the second using WWM and 0.2\% masking rate. Results are presented in Table 3.



\begin{table}[!htbp]
\begin{adjustbox}{width=\columnwidth,center}
\begin{tabular}{ccccc}
\hline
\multicolumn{1}{l}{}              & \begin{tabular}[c]{@{}c@{}}BM \\ from scratch\end{tabular} & \begin{tabular}[c]{@{}c@{}}BM \\ continued\end{tabular} & \begin{tabular}[c]{@{}c@{}}BM \\ averaged \end{tabular} & \begin{tabular}[c]{@{}c@{}}BM \\ contextualized\end{tabular} \\ \hline
\multicolumn{1}{c}{BC5-chem}     & 92.51                                                                    & \textbf{93.7}                                                        & 92.93                                                                                 & { 93.52}                                                                \\
\multicolumn{1}{c}{BC5-disease}  & 80.66                                                                    & \textbf{83.92}                                                        & 81.83                                                                                 & { 83.41}                                                                \\

\multicolumn{1}{c}{NCBI-disease} & 86.34                                                                    & 86.67                                                                 & 87.38                                                                                 & \textbf{88.44}                                                             \\
\multicolumn{1}{c}{BC2GM}        & 81.4                                                                     & \textbf{83.78}                                                        & 82.82                                                                                 & {\ 83.53}                                                                \\
\multicolumn{1}{c}{JNLPBA}       & { 76.14}                                                                    & 77.59                                                        & 75.89                                                                                 & \textbf{78.71}                                                                      \\ \hline
\end{tabular}
\end{adjustbox}
\caption{Performance (F1) of test models on five NER tasks.}
\label{tab:my-table}
\end{table}

\begin{table}[]
\begin{adjustbox}{width=\columnwidth,center}
\begin{tabular}{cccc}
\hline
\multicolumn{1}{l}{}                       & \textbf{\begin{tabular}[c]{@{}c@{}} contextualized\\ + MLM\\ + 0.15 mr\end{tabular}} & \textbf{\begin{tabular}[c]{@{}c@{}}contextualized\\ + WWM \\+ 0.15 mr\end{tabular}} & \textbf{\begin{tabular}[c]{@{}c@{}}contextualized\\  + WWM  \\+  0.2 mr\end{tabular}} \\ \hline
\multicolumn{1}{c}{\textbf{BC5-chem}}     & 93.52                                                                                                                           & {\ul 93.77}                                                                                                           & \textbf{93.96}                                                                                                       \\
\multicolumn{1}{c}{\textbf{BC5-disease}}  & 83.41                                                                                                                           & {\ul 84.02}                                                                                                           & \textbf{84.28}                                                                                                       \\

\multicolumn{1}{c}{\textbf{NCBI-disease}} & {\ul 88.44}                                                                                                                     & 87.41                                                                                                                 & \textbf{88.85}                                                                                                       \\
\multicolumn{1}{c}{\textbf{BC2GM}}        &  83.53                                                                                                                     & {\ul 83.76}                                                                                                                 & \textbf{83.89}                                                                                                       \\
\multicolumn{1}{c}{\textbf{JNLPBA}}       & {\ul 78.71}                                                                                                                     & 78.35                                                                                                                 & \textbf{78.8}                                                                                                       \\ \hline
\end{tabular}
\end{adjustbox}
\caption{Performance (F1) of models pre-trained with contextualized embeddings and different pre-training techniques. The scores of the best-performing models are in bold, and underlined are the scores of the second-best-performing models.}
\label{tab:my-table}
\end{table}

\section{Analysis}

As shown in Table 2, \textbf{BM continued} shows superior performance on three datasets. The worst-performing model is \textbf{BM from scratch}. Such performance is expected due to the experimental set-up and pre-training for a shorter period. The performance of PubMedBERT (Table 6) suggests that pre-training from scratch with a domain-specific vocabulary can outperform continued pre-training but would require more extended training to reach this point. \textbf{BM averaged} performs better than the model pre-trained from scratch, demonstrating that averaging weights of subtokens to initialize weights for new tokens gives an initial boost for the model to learn. However, the second best-performing model is \textbf{BM contextualized} or \textbf{BIOptimus 0.1}. 
It outperforms \textbf{BM averaged} and \textbf{BM from scratch} on all datasets and \textbf{BM continued} on two datasets.

Figure 3 shows the models' performance on the dev set after each epoch. We can see that \textbf{BM continued} learns quickly and performs better on JNLPBA and BC5-chem datasets. However, the model shows the lowest learning curve on the NCBI-disease dataset, which the absence of specialized vocabulary can explain. Figure 3 also illustrates the importance of domain-specific vocabulary, and we can see that all the models with specialized vocabulary learn quickly and perform relatively well. Overall, introducing domain-specific vocabulary is beneficial. However, it is less efficient to pre-train a new model entirely from scratch, initializing all weights randomly. In the case of BIOptimus 0.1, adding new domain-specific tokens alongside the corresponding weights initialized in a way that leverages the contextualizing ability of transformers gives noticeable gains.
Additionally, it is important to compare how biomedical LMs respond to pre-training on WWM against pre-training with masking only tokens/subtokens. Table 3 shows that it is not precisely the case that the WWM pre-training performs better across the board (\citealp{dai-etal-2022-whole}; \citealp{32_gu_2021}). 
Pre-training with masking only tokens performs better than WWM on two datasets out of five with a masking rate of
0.15 (underlined scores). One of the explanations is that individual subtokens often coincide with morphological components of the words, and learning the meanings of these components separately may be beneficial for the biomedical LM \cite{hofmann-etal-2020-dagobert}. Moreover, Table 3 demonstrates that increasing the masking rate contributes to a tapered performance enhancement.

\begin{figure*}[h]
\centering
\includegraphics[width=\textwidth]{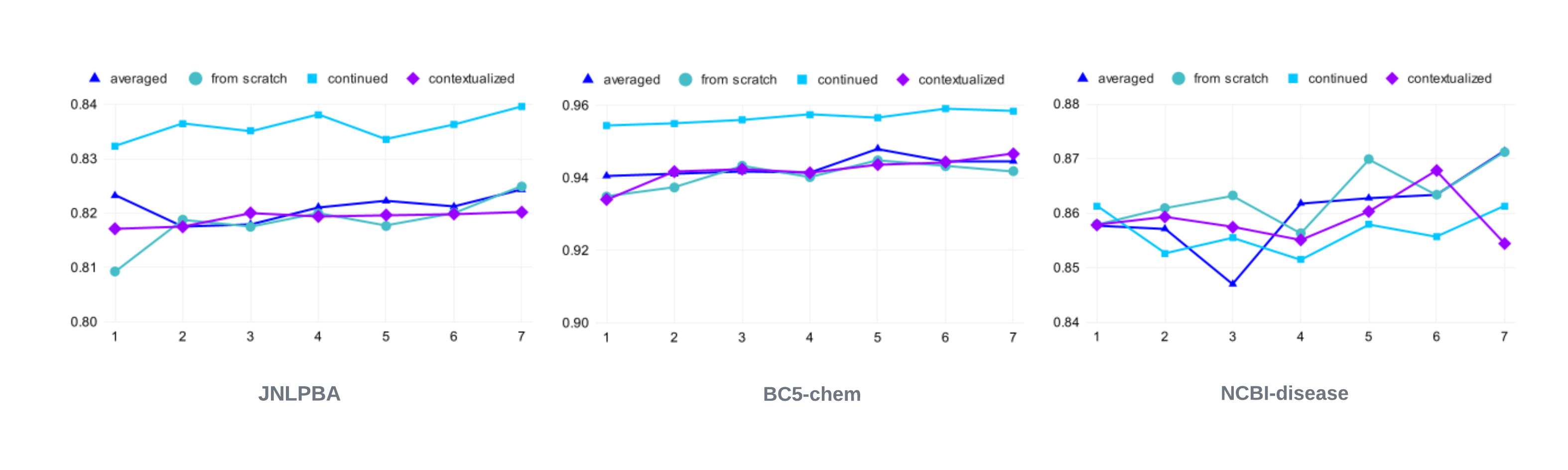}
\caption{Effects of pre-training epochs on performance (F1) of four test models on the development set of JNLPBA, BC5-chem, and NCBI-disease datasets.}
\end{figure*}

\section{Curriculum Learning}
\subsection{Motivation and Method}
Analysis of our experiments motivates us to hypothesize that different training techniques like masking rate and masking strategies might help broaden the model’s experience \cite{mitchell1997machine} and gain more diversified knowledge of textual input. In addition, during the experimental stage, we’ve observed that pre-training with specific techniques like WWM and increased masking rate slow down the training process if introduced right from the beginning due to increased task complexity. That brings us to the idea that introducing more complex tasks gradually, using CL’s easy-to-hard strategy, assists in guiding the model’s learning process more smoothly \cite{bengio_2009}.
Using the results of our experiments, we implement a CL method for pre-training a biomedical LM. In Masked Language Modeling (MLM), the objective is to predict the masked token based on the surrounding context:
\begin{equation}
    \begin{split}
        L_{MLM} = - \sum_{i}\log P_\Theta (\omega_i|\tilde{\omega_i}) \\ = - \sum_{i}\log\frac{\exp{(E(\omega_i)^{\top}\tilde{h_i})}}{\sum_{j=1}^{\left |V  \right |} \exp{(E(\omega_i)^{\top}\tilde{h_i})}}
    \end{split}
\end{equation}

It remains a challenging task to measure the difficulty of a task or a training sample in CL. It is primarily an issue in the case of pre-training masked language models. There are a few newly proposed approaches to tackle this challenge. \citet{nagatsuka-etal-2021-pre} pre-trains RoBERTa by increasing the block size of input text. \citet{lee-etal-2022-efficient-pre}  proposed a curriculum based on masking easy-to-predict tokens first.  We formulate our curriculum strategy from the perspective of the complexity of the prediction task. Predicting whole words is more complex than predicting just tokens which may be a part of the word, giving the LM more hints from the surrounding context (\citealp{32_gu_2021}; \citealp{Cui_2021}). Increasing the masking rate makes prediction more challenging \cite{wettig2023mask} since less context is available for prediction.

We use a pre-trained model of the same architecture and number of parameters to measure the prediction task difficulty and evaluate its performance as MLM (\citealp{dudy-bedrick-2020-words}; \citealp{12_liu_2019}). We use a corpus from a different domain to account for domain shift. In our case, we evaluate the performance of the BERT-base-uncased\footnote{\url{https://huggingface.co/bert-base-uncased}} on the RealNews dataset \cite{2_raffel_2019}. The evaluation results are presented in Figure 4. Based on this evaluation, we divide our curriculum into four phases (see Table 4).

We start with pre-training vanilla MLM (masking random tokens with a masking rate of 0.15). As we can see from Figure 4, this task is the easiest to handle for MLM. This task also ensures the model learns the subtokens that constitute complex biomedical terminology.  At the next stage, we increase the complexity of the prediction task and pre-train predicting the whole words to teach the model to combine separate subtokens into whole words. In the third phase, we raise the task complexity to one more level by increasing the masking rate to 0.2 in predicting the whole words. \citet{wettig2023mask} observed that the corruption rule "80-10-10" hurts the performance for some downstream tasks and suggested using only [MASK] without corruption strategy for MLM pre-training. In our training experiments, we observe that prediction without corruption of tokens makes the prediction task even more complex and slows down the learning process; thus, we add it as an additional curriculum phase at the end. 

We track and visualize contextualized word representations with different frequencies to observe how pre-training phases evolve over time and their impact on the performance and quality of pre-trained embeddings. Figure 5 shows that low-frequency words form a separate cluster from high-frequency words, which is more evident after the first stage. During pre-training, the gap decreases; however, only after the fourth phase, when the corruption strategy is removed, do the clusters join closer together. It is plausible that tokens’ prediction with a corruption strategy plays a role in degenerating word embeddings, and it may explain why it hurts the performance in some downstream tasks \cite{wettig2023mask}. We leave further experiments on this subject for future research.
\begin{table}[]
\begin{adjustbox}{width=\columnwidth,center}
\begin{tabular}{ccccc}
\hline
\multicolumn{1}{l}{v.}               & \textbf{Phases} & \textbf{\thead{Masking\\strategy}} & \textbf{\thead{Masking\\rate}} & \textbf{\thead{Corruption\\strategy}} \\ \hline
\multicolumn{1}{c}{0.1} & phase 1                           & tokens                    & 0.15                  & with corruption              \\
\multicolumn{1}{c}{0.2} & phase 2                           & \textbf{WWM}              & 0.15                  & with corruption              \\
\multicolumn{1}{c}{0.3} & phase 3                           & \textbf{WWM}              & \textbf{0.2}          & with corruption              \\
\multicolumn{1}{c}{0.4} & phase 4                           & \textbf{WWM}              & \textbf{0.2}          & \textbf{no corruption}       \\ \hline
\end{tabular}
\end{adjustbox}
\caption{Stages of the CL method with increasing task complexity.}
\label{tab:my-table}
\end{table}

\begin{figure}[]
\centering
\includegraphics[width=1.0\columnwidth]{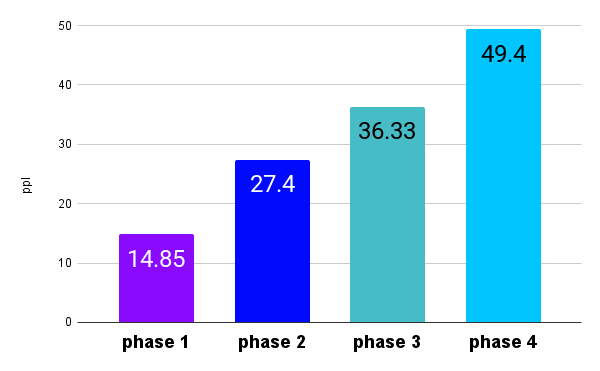}
\caption{Evaluation of the performance of the BERT-base-uncased on the RealNews corpus (perplexity). We average the scores across text files randomly sampled from the corpus. We take ten samples, each of size of around 12M tokens.}
\end{figure}

\begin{figure}[h]
\centering
\includegraphics[width=\columnwidth]{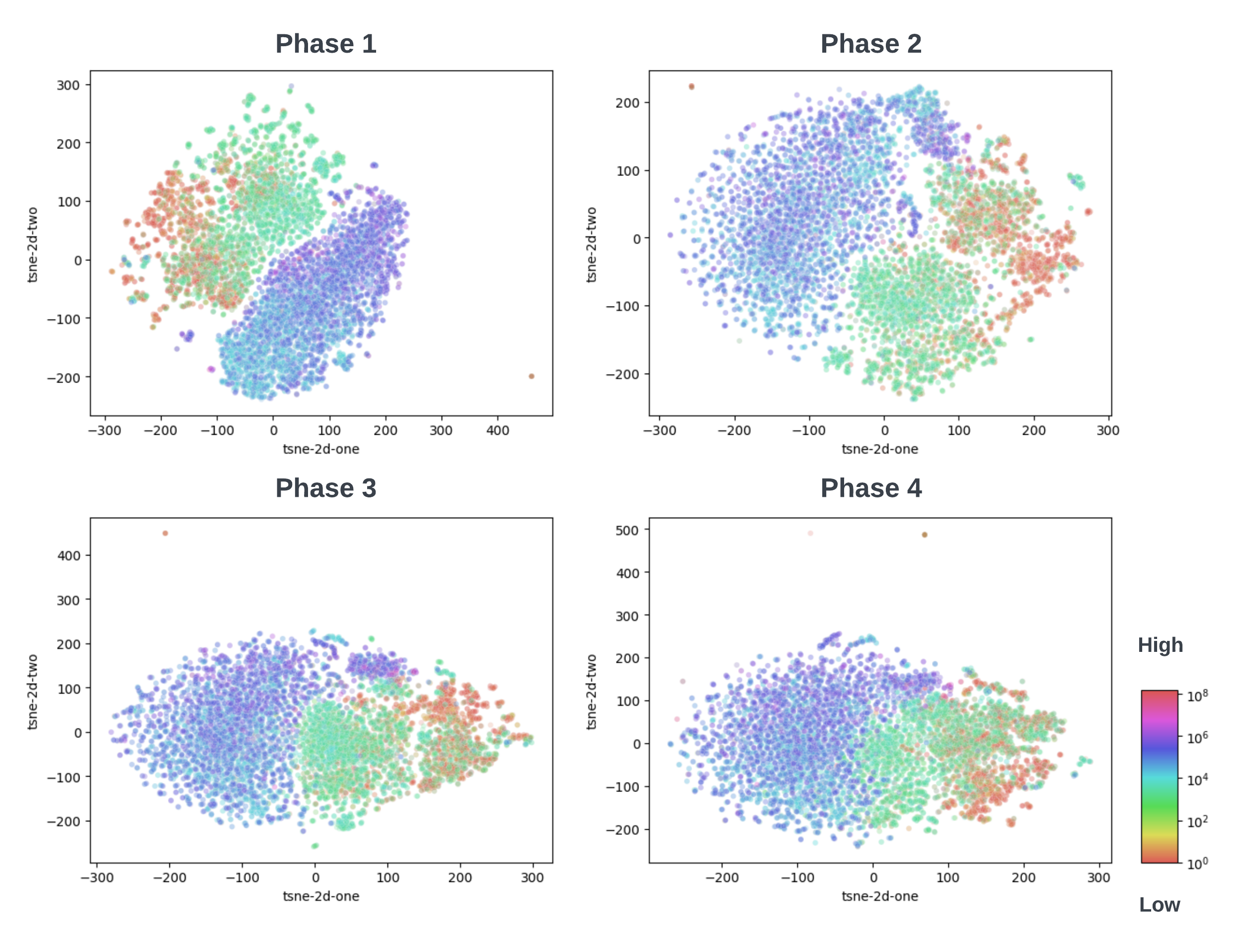}
\caption{A t-SNE visualization of word embeddings during four pre-training phases with the CL method. \textcolor{colorhigh}{"High"} on the color bar corresponds to high frequencies, and \textcolor{colorlow}{"Low"} to low frequencies.}
\end{figure}

\begin{table*}[th!]
\begin{tabular}{cccccc}

\hline
 \multicolumn{1}{l}{v.}               & BC5-chem & BC5-disease & NCBI-disease & BC2GM & JNLPBA \\ \hline
 \multicolumn{1}{c|}{0.1} & \multicolumn{1}{c|}{93.52}    & \multicolumn{1}{c|}{83.41}       & \multicolumn{1}{c|}{88.44}        & \multicolumn{1}{c|}{83.53} & 78.71  \\
 \multicolumn{1}{c|}{0.2} & \multicolumn{1}{c|}{93.87}    & \multicolumn{1}{c|}{83.43}       & \multicolumn{1}{c|}{87.99}        & \multicolumn{1}{c|}{84.16} & 79.12  \\ \multicolumn{1}{c|}{0.3} & \multicolumn{1}{c|}{93.7}     & \multicolumn{1}{c|}{85.06}       & \multicolumn{1}{c|}{88.17}        & \multicolumn{1}{c|}{84.54} & 79.28  \\
 \multicolumn{1}{c|}{0.4} & \multicolumn{1}{c|}{94.1}     & \multicolumn{1}{c|}{84.98}       & \multicolumn{1}{c|}{89.54}        & \multicolumn{1}{c|}{85.25} & 79.46  \\ 
 \multicolumn{1}{c|}{}               & \multicolumn{1}{c|}{\includegraphics[width=0.16\linewidth]{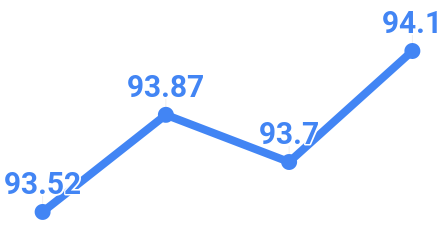}}    & \multicolumn{1}{c|}{\includegraphics[width=0.16\linewidth]{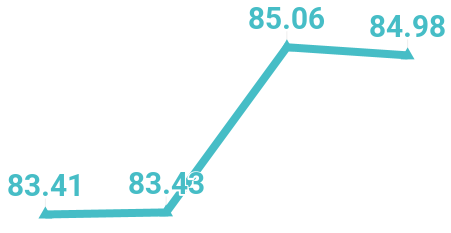}}        & \multicolumn{1}{c|}{\includegraphics[width=0.16\linewidth]{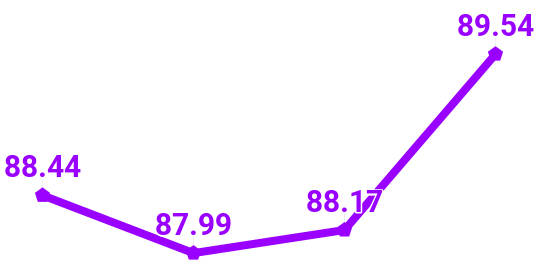}}         & \multicolumn{1}{c|}{\includegraphics[width=0.16\linewidth]{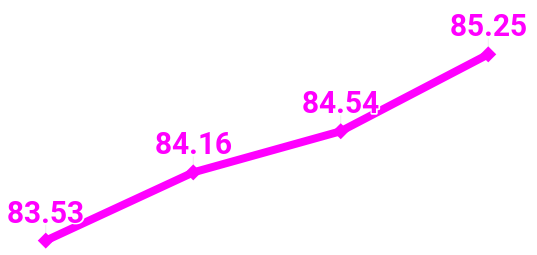}}  & \includegraphics[width=0.16\linewidth]{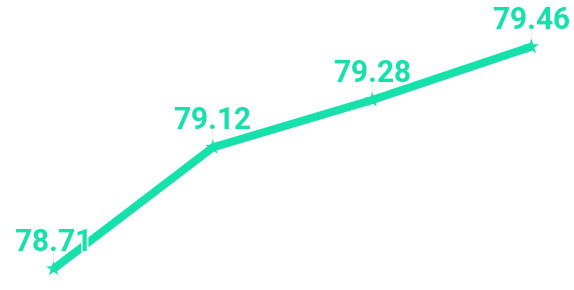}   \\ \hline
\end{tabular}
\caption{Evaluation of each CL phase on NER datasets.}
\label{tab:my-table}
\end{table*}

\subsection{Experimental Setting}
\textbf{Data.} We use the same PubMed Corpus and NER datasets described in Section 5.  

\textbf{Implementation.} The model’s weights are initialized using our contextualized weight distillation approach, which helps speed up pre-training.  We pre-train with the same hyperparameters presented in Appendix A and increase the number of optimization steps for one epoch for each subsequent phase to account for the complexity of the prediction task. The initial learning rate is set at 1e-4, and when restarting the training at the next phase, we decrease the learning rate following the learning rate scheduler.

\textbf{Models and Baselines.} We track the progress of the CL method by evaluating models after each phase (see Table 5). Each phase's resulting model is considered an independent model and brings competitive results right from the first phase. Furthermore, we compare our models with BioBERT \cite{6_lee_2019}, PubMedBERT \cite{32_gu_2021}, and BioLinkBERT-base \cite{yasunaga-etal-2022-linkbert} (Table 6).

\begin{table}[th!]
\newcommand{\wrap}[1]{\parbox{.20\linewidth}{\vspace{1.5mm}#1\vspace{1mm}}}
\begin{adjustbox}{width=\columnwidth,center}
\begin{tabular}{ccccc}
\hline
             & \wrap{\centering\textbf{Bio-\\BERT}} & \wrap{\centering\textbf{PubMed-\\BERT}} & \wrap{\centering\textbf{BioLink-\\BERT}} & \wrap{\centering\ \textbf{BIOptimus\\ 0.4}} \\ \hline
\textbf{BC5-chem}     & 92.85            & 93.33               & \uline{93.75}                                 & \textbf{94.1}          \\
\textbf{BC5-disease}  & 84.7             & \uline{85.62}               & \textbf{86.1}                                & 84.98                  \\
\textbf{NCBI-disease} & \uline{89.13}            & 87.82               & 88.18                                        & \textbf{89.54}         \\
\textbf{BC2GM}        & 83.82            & 84.52               & \uline{84.9}                                   & \textbf{85.25}         \\
\textbf{JNLPBA}       & 78.55            & \textbf{80.06}      & 79.03                                        & \uline{79.46}                  \\ \hline
\end{tabular}
\end{adjustbox}
\caption{Comparison of pre-trained language models on five NER datasets. The scores of the best-performing models are in bold, and underlined are the scores of the second-best-performing models. BIOptimus 0.4 sets a new state-of-the-art on BC5-chem, NCBI-disease, and BC2GM datasets.}
\label{tab:my-table}
\end{table}

\subsection{Results and Discussion}
\textbf{Masking strategy (from phase 1 to phase 2)} (see Table 5) gives noticeable gains in performance for BC5-chem, BC2GM, and JNLPBA datasets, there is only a slight improvement on the BC5-disease dataset, and the NCBI-disease dataset responded quite poorly to this transition. 
\textbf{Increasing the masking rate (from phase 2 to phase 3)} helps to advance performance on all datasets except BC5-chem, while BC5-disease shows a considerable gain from increasing the masking rate.
\textbf{Removing the corruption strategy "80-10-10" (from phase 3 to phase 4)} is generally beneficial for all datasets, with considerable gains for BC5-chem and NCBI-disease and a slight drop for BC5-disease.
BC2GM and JNLPBA datasets respond with stable improvement in all pre-training phases of the CL method. BC5-chem, BC5-disease, and NCBI-disease datasets exhibit more diverse responses to changes in curriculum phases.

\subsection{Ablation Study}
In this section, we conduct an ablation study to assess the effect of the proposed CL method. To measure the impact of pre-training with our CL method, we pre-train a model in a continued setting with the contextualized weight distillation method, using vanilla MLM (with a masking rate of 0.15, applying a corruption rule “80-10-10”) but without CL. Essentially the model is BIOptimus 0.1 pre-trained for the same number of optimization steps as BIOptimus 0.4. The performance of this model is presented in Table 7 (“No CL”). Removing the CL method hurts downstream performance. The drop occurs with all NER datasets and is more apparent with BC5-disease, NCBI-disease and BC2GM datasets. This suggests that pre-training with the CL method helps boost biomedical LM's performance on NER task.

\begin{table}[h]
\begin{adjustbox}{width=\columnwidth,center}
\begin{tabular}{cccccc}
\hline
\multicolumn{1}{l}{}                        & \textbf{BC5-chem} & \textbf{BC5-disease} & \textbf{NCBI-disease} & \textbf{BC2GM} & \textbf{JNLPBA} \\ \hline
\multicolumn{1}{c}{\textbf{BIOptimus 0.4}} & \textbf{94.1}     & \textbf{84.98}       & \textbf{89.54}        & \textbf{85.25} & \textbf{79.46}                          \\
\multicolumn{1}{c}{\textbf{No CL}}         & 93.86             & 84.29                & 88.92                 & 84.53          & 79.07    \\
\hline
\end{tabular}
\end{adjustbox}
\caption{Ablation study on the CL method.}
\label{tab:my-table}
\end{table}

\section{Conclusion}
This paper presented a new method to initialize tokens’ weight for new biomedical vocabulary when pre-training from the existing checkpoint (continued approach). We also compared this method of token weight initialization with other pre-training methods (see Table 2 and Table 3). This method showed considerable gains in speeding up the pre-training phase and improving performance on NER. Comparing pre-training techniques showed that WWM is not the best-performing approach for all NER tasks, and masking only tokens/subtokens shows competitive performance. Increasing the masking rate and removing the corruption strategy are generally beneficial techniques for pre-training biomedical LM. Finally, we introduced the CL method based on the task complexity to pre-train LMs. The “easy-to-hard” CL method introduces the biomedical LM to a broader scope of language experience, speeds up pre-training, and enhances performance on downstream tasks like NER. It is important to highlight that our model BIOptimus 0.4 achieves high performance with the pre-training time reduced by at least half, proving the pre-training approach's efficiency.

\section*{Acknowledgment}
We thank Sebastian Hurubaru and our anonymous reviewers for their valuable feedback. This work was supported by tnsr capital.

\section*{Limitations}
(1) To be able to present a rigorous comparison and analysis of pre-training an optimal biomedical LM, we focused on running extensive evaluation on five NER tasks from the BLURB benchmark\footnote{\url{https://microsoft.github.io/BLURB/}}. We do not evaluate the performance of our models on other downstream tasks in the framework of this paper and leave it for future work. (2) While we performed additional experiments to explain the reason for the performance drop on some tasks when implementing the corruption strategy "80-10-10," one plausible explanation is that it might increase the degree of anisotropy in the contextualized word representations. Separate work is needed to search for how the corruption strategy might cause a drop in performance on some downstream tasks. (3)  Due to the expensive nature of pre-training experiments, we were not able to experiment with all possible combinations of pre-training methods and techniques. (4) Our models were pre-trained on the English language only.

\section*{Ethics Statement}
While our research does not directly introduce any social or ethical bias nor amplify the bias in the data, we inherit a considerable amount of the underlying limitations of LMs. LM pre-training is computationally expensive and causes environmental damage. Our research focuses on overall computing resource efficiency and thus has a marginal environmental footprint.

\bibliography{anthology,custom}
\bibliographystyle{acl_natbib}

\vspace{\fill}
\pagebreak
\vspace{\fill}
\pagebreak

\appendix

\section{Appendix}
\label{sec:appendix}
\subsection{Datasets}

\begin{figure}[h!]
\includegraphics[width=\columnwidth]{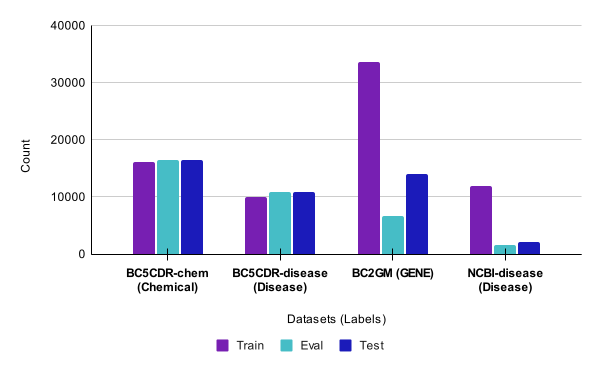}
\caption{Label distribution of BC5CDR-chem, BC5CDR-disease, BC2GM, NCBI-disease datasets.}
\end{figure}

\vspace{\fill}

\subsection{Computing Infrastructure}
\begin{table}[h!]
\begin{adjustbox}{width=\columnwidth,center}
\begin{tabular}{ccc}
\hline
\textbf{Computing Infrastructure}    & \multicolumn{2}{c}{8 x A100 GPUs}    \\ \hline
\multicolumn{2}{c}{\textbf{Hyperparameter}} & \textbf{Assignment}                     \\ \hline
\multicolumn{2}{c}{number of epochs}        & 7-34 \\ \hline
\multicolumn{2}{c}{batch size}              & 256                             \\ \hline
\multicolumn{2}{c}{maximum learning rate}   & 0.0005                        \\ \hline
\multicolumn{2}{c}{learning rate optimizer} & Adam                                    \\ \hline
\multicolumn{2}{c}{learning rate scheduler} & None or Warmup linear                   \\ \hline
\multicolumn{2}{c}{Weight decay}            & 0.01                                    \\ \hline
\multicolumn{2}{c}{Warmup proportion}       & 0.06                                    \\ \hline
\multicolumn{2}{c}{learning rate decay}     & linear                                  \\ \hline
\end{tabular}
\end{adjustbox}
\caption{Hyperparameters for pre-training biomedical LMs.}
\label{tab:appendix-table-a}
\end{table}

\begin{table}[h!]
\begin{adjustbox}{width=\columnwidth,center}
\begin{tabular}{ccc}
\hline
\textbf{Computing Infrastructure} & \multicolumn{2}{c}{2 x NVIDIA RTX 3090 GPU} \\ \hline

\multicolumn{2}{c}{\textbf{Hyperparameter}}     & \textbf{Assignment}           \\ \hline
\multicolumn{2}{c}{number of epochs}            & 5-11                             \\ \hline
\multicolumn{2}{c}{batch size}                  & 4, 8, 16                            \\ \hline
\multicolumn{2}{c}{learning rate}               & 1e-5, 2e-5                          \\ \hline
\multicolumn{2}{c}{dropout}                     & 0.1                           \\ \hline
\end{tabular}
\end{adjustbox}
\caption{Hyperparameters for fine-tuning on NER datasets.}
\label{tab:appendix-table-b}
\end{table}

\end{document}